%% file: main.tex
\documentclass[fullpaper]{nldl}

\renewcommand{\DOI}{12345}

\usepackage[utf8]{inputenc}
\usepackage{url}
\usepackage{graphicx}
\usepackage{authblk}

\usepackage{algorithm}
\usepackage{algorithmic}

\usepackage[square,numbers]{natbib}

\usepackage{hyperref}

\usepackage{comment}
\usepackage{amsmath}
\usepackage{booktabs}
\usepackage{makecell}
\usepackage{multirow}
\usepackage{subcaption}
\usepackage{layouts}
\usepackage{cancel}

\title{Neural machine translation for automated feedback on children's early-stage writing}
\author[]{Jonas Vestergaard Jensen\thanks{Corresponding Author: jovje@dtu.dk}}
\author[]{Mikkel Jordahn}
\author[]{Michael Riis Andersen}
\affil[]{Technical University of Denmark}

\date{\vspace{-5ex}}

\begin{document}
\nldlmaketitle
\begin{abstract}  
    In this work, we address the problem of assessing and constructing feedback for early-stage writing automatically using machine learning. Early-stage writing is typically vastly different from conventional writing due to phonetic spelling and lack of proper grammar, punctuation, spacing etc. Consequently, early-stage writing is highly non-trivial to analyze using common linguistic metrics. We propose to use sequence-to-sequence models for "translating" early-stage writing by students into "conventional" writing, which allows the translated text to be analyzed using linguistic metrics. Furthermore, we propose a novel robust likelihood to mitigate the effect of noise in the dataset. We investigate the proposed methods using a set of numerical experiments and demonstrate that the conventional text can be predicted with high accuracy.
\end{abstract}

\section{Introduction}
\label{sec:introduction}
Learning to write is extremely important for both educational and communication purposes. Specific and frequent formative feedback can improve learning, but it is a time-consuming process, especially in a class-room setting \cite{SaliuAbdulahi2017TeachersF}. In this work, we study the problem of using machine learning to assist elementary school teachers in assessing and constructing formative feedback for early-stage writing.

Children's early writing can be studied and quantified in several ways ranging from simple count statistics to more sophisticated linguistic metrics and such metrics can be tracked over time to assess and facilitate learning \cite{BundsgaardKabelBremholm2022}.  However, emergent writing is often characterized by phonetic spelling as well as lack of proper grammar, spacing, and punctuation etc.\ \cite{KabelBremholmBundsgaard2022}, which makes automatic and quantitative analysis highly non-trivial.

To address this problem, we propose to use neural machine translation models to "translate" the early-writing of a student to the equivalent "conventional" writing. This makes it possible to compare and quantify the difference between student texts and corresponding conventional texts and to evaluate the texts using linguistic metrics of interest. That is, we model the writing produced by students as noisy observations of the corresponding conventional writing and thus aim to "de-noise" the student texts using sequence-to-sequence models.

Due to the current success of Transformer-based models in many natural language processing (NLP) applications \cite{Transformers}, we employ the so-called BART architecture, which is a Transformer-based denoising autoencoder for sequence-to-sequence problems \cite{lewis2020bart}. For a review on recent sequence-to-sequence methods for neural machine translation, see e.g.\ \citet{stahlberg2020neural} or \citet{tan2020neural}. Furthermore, we refer to the work by \citet{Ramesh2022} and \citet{Beigman2020} for recent systematic reviews of automated essay scoring.

We train and evaluate the model on a dataset collected using a digital learning platform\footnote{www.writereader.com}.
The dataset consists of $N = 36,610$ pieces of early writing produced by students and the corresponding conventional texts produced by teachers after interacting with the students, i.e.\ the dataset is $\mathcal{D} = \left\lbrace (x_n, y_n) \right\rbrace_{n=1}^N$, where $x_n$ and $y_n$ are the $n$'th student and teacher texts, respectively. The following three pieces of texts are examples of student writing: 'We lern ubut  eath in sins.', 'thedinousouisrune', or 'ledkos boo fune thengs' giving rise to the following conventional texts 'We learn about Earth in Science.', 'The dinosaur runs', and 'Leprechauns do funny things.', respectively. 

The dataset contains a significant amount of noise caused by students or teachers using the learning platform in unintended ways. In approximately 25\% of the data, there is no relationship between the texts in the pair $(x_n, y_n)$. For example, a student wrote 'norah loves peas!' and the corresponding teacher text is 'Elephants are big.'. To address this problem, we propose a novel robust likelihood for sequence-to-sequence modelling. 

To evaluate the proposed methods, we investigate and report how accurately a teacher text $y_*$ can be predicted from the student text $x_*$ given a training set $\mathcal{D}$. We also consider the task of estimating two readability metrics (Flesch–Kincaid \cite{kincaid1975derivation} and LIX \cite{bjornsson1968lasbarhet}) from the students texts.

The main contributions of this paper are 1) to demonstrate that the student texts can be de-noised with high accuracy, 2) to demonstrate that translation of the student texts to conventional writing significantly improves the accuracy of the estimated linguistic metrics, 3) introduction and evaluation of a novel likelihood for noisy sequence-to-sequence data, and 4) investigation of whether the quality of a translation can be assessed through its average likelihood.

\section{Methods}
\label{sec:methods}

\subsection{Sequence-to-sequence modelling}
We frame the problem as a sequence-to-sequence problem where the goal is to estimate the teacher text $y_n$ given a student text $x_n$.
In other words, given a training set $\mathcal{D} = \left\lbrace (x_n, y_n) \right\rbrace_{n=1}^N$, our goal is to estimate the distribution $p(y_*|x_*, \mathcal{D})$ for some new student text $x_*$. Formally, each sequence $x_n = (x_{n, 1}, x_{n, 2}, \dots, x_{n, N_x})$ consists of an ordered list of tokens from a fixed vocabulary $\mathcal{A}$, i.e.\ $x_{n,i} \in \mathcal{A}$, where  $N_{x_n}$ denotes the length of $x_n$. We use $K$ to denote the size of the vocubulary, i.e.\ $K = |\mathcal{A}|$. Furthermore, we use the notation $x_{n, 1:j}$ to denote the first $j$ tokens of $x_n$, i.e.\ the subsequence $x_{n, 1:j} = (x_{n, 1}, x_{n, 2}, \dots, x_{n, j})$. We use the same notation for $y_n$. Finally, we sometimes omit the data index $n$ and simply write $y_i$ for the $i$'th token in $y$ and $y_{1:j}$ for the first $j$ tokens in $y$.

\paragraph{Model archictecture} We use the BART sequence-to-sequence architecture \cite{lewis2020bart} to model $p(y_n|x_n)$.
BART uses an encoder-decoder architecture with a bidirectional Transformer model \cite{devlin-etal-2019-bert} as the encoder and an autoregressive Transformer model \cite{Radford2018ImprovingLU} as the decoder yielding the following likelihood for the $n$'th observation
\begin{align} \label{eq:BART}
    p(y_n|x_n) = \prod_{i=1}^{N_{y}} p(y_{n,i}\vert x_n, y_{n,1:i-1}).
\end{align}
After training the model, we can predict $y$ using greedy search as follows
\begin{align} \label{eq:predict}
    \hat{y}_{n,i} = \arg\max\limits_{k} p(y_{n,i}=k\vert x_n, \hat{y}_{n,1:i-1}),
\end{align}
where $\hat{y}_{n,i}$ denotes the prediction for the $i$'th token in the $n$'th example.

\paragraph{Loss function and label smoothing} Due to the autoregressive nature of the model, predicting each toking in $y_n$ is a multi-class classification problem, and hence, the cross-entropy loss function is a natural choice
\begin{align*}
    \ell(x,y) = -\sum_{i=1}^{N_y}\sum_{k=1}^K q(y_i=k) \log p(y_i=k \vert x, y_{1:i-1}),
\end{align*}
where $q(y_i=k) = \delta_{k,y_i}$ is a Kroncker's delta function such that $\delta_{k, y_i} = 1$ if $y_i = k$ and $0$ otherwise. 

We employ label smoothing for regularization like \citet{lewis2020bart}. That is, $q(y_i)$ is replaced with a mixture between $q(y_i)$ and a uniform distribution over the vocabulary $q'(y_i=k) = (1-\epsilon)\delta_{k,y_i} + \epsilon \frac{1}{K}$ \cite{labelsmoothing}, where $K$ is the size of the vocabulary and $\epsilon \in \left[0, 1\right]$ is the smoothing parameter.
\subsection{Robust likelihood for noisy data} 
As mentioned in the introduction, a significant proportion of the observations in the dataset is contaminated with noise.
It has been shown that noise in training datasets can dramatically decrease prediction performance \cite{Gupta2019DealingWN}. \citet{robustlikelihood} proposed a likelihood for multi-class classification that accounts for labelling errors in the dataset. Inspired by this work, we propose a novel likelihood for robust sequence-to-sequence modelling to mitigate the effect of the noise in the data. We construct the likelihood using the following generative process.
For each example, we introduce a latent binary variable, $\theta_n \in \left\lbrace 0, 1 \right\rbrace$, indicating whether the corresponding target sequence $y_n$ is noisy ($\theta_n=1$) or not ($\theta_n=0$). If $y_n$ is noise-free, i.e.\ $\theta_n = 0$, then we model $y_n$ conditionally as $p(y_n|x_n, \theta_n=0) = p_\text{BART}(y_n|x_n)$ from eq.\ \eqref{eq:BART}. On the other hand, if $y_n$ is noisy (i.e.\ $\theta_n=1$), then we assume $p(y_n|x_n, \theta_n=1) = p_{\text{LM}}(y_n)$, where $p_\text{LM}(y_n)$ is a language model independent of $x_n$. Thus, the robust likelihood for the $n$'th example is
\begin{align*}
        p(y_n|x_n, \theta_n) = \left[p_{\mathrm{BART}}(y_n | x_n)\right]^{1-\theta_n} \left[p_{\mathrm{LM}}(y_n)\right]^{\theta_n}.
\end{align*}
Imposing i.i.d.\ Bernoulli distributions on the latent indicator variables, i.e.\ $\theta_n \sim \text{Ber}(\alpha)$, and marginalizing yields the following robust likelihood
\begin{align}
\label{eq:rllh}
    p(y_n|x_n) = (1-\alpha) p_{\mathrm{BART}}(y_n | x_n) + \alpha p_{\mathrm{LM}}(y_n),
\end{align}
where $\alpha \in \left[0, 1\right]$ controls the rate of noisy examples. In the special case, where $\alpha \rightarrow 0$, we recover the classic likelihood from eq.\ \eqref{eq:BART}. On the other hand, when $\alpha \rightarrow 1$ the model becomes a language model independent of $x_n$. The language model can range from a simple uniform distribution to an $n$-gram model to a complex neural language model.

For a pair $(x_n, y_n)$ with no relationship between $x_n$ and $y_n$, we expect that $p_{LM}(y_n) > p_{\mathrm{BART}}(y_n | x_n)$ on average and therefore a lesser contribution to the loss. We confirmed this behavior empirically.

\subsection{Calibration and decision-making}
Reconstruction of the teacher text $y_n$ from $x_n$ can be an ill-posed problem in the sense that $y_n$ is not always uniquely determined from $x_n$. For example, if students are really early in their writing development or not focused on the writing task, then $x_n$ may contain very little information about what the student intended to write and in these cases, it is not possible to predict $y_n$ from $x_n$ alone. For example, the conventional writing of the student text 'de bea r pae' is 'These bears are playing.', which is not obvious. However, using eq.\ \eqref{eq:predict} always leads to a prediction.

To be able to reject predictions for such examples, we investigate to what degree the average log-likelihood, i.e.
\begin{align*}
    C(\hat{y}| x) = \frac{1}{N_{\hat{y}}}\sum_{i=1}^{N_{\hat{y}}} \log p(\hat{y}_i \vert x, \hat{y}_{1:i-1})
\end{align*}
of the translation $\hat{y}$ of $x$ reflects the quality of the predicted text $\hat{y}$. Intuitively, a translation $\hat{y}$ with a low likelihood should be a poor translation of the student text, and therefore, unsuitable to use as a basis for evaluating downstream linguistic metrics.

\paragraph{Model calibration} It is well-known that calibrated uncertainties are required for optimal decision-making \cite{berger2013statistical} and the argument above does indeed assume that the models are calibrated \cite{RevisitngCalNN}. However, neural networks can be overconfident in their predictions \citep{RevisitngCalNN, calibPreTrainedTrans, guo2017calibration}.
Therefore, we consider two methods for improving model calibration: re-calibration via temperature scaling \cite{guo2017calibration} and Deep Ensembles (DE) \cite{DeepEnsemblesProposed} and investigate whether they improve the calibration of the model, and subsequently, lead to better decision-making. 

\paragraph{Temperature scaling} In general, the probabilities $p(y_i=k\vert x, y_{1:i-1})$ are computed by feeding the network outputs, $z_{i,k}$, through the softmax function, i.e.\ $p(y_i = k \vert x, y_{1:i-1}) = \frac{\exp z_{i,k}}{\sum_{k=1}^K \exp z_{i,k}}$. In temperature scaling, the logits $z_{i,k}$ are simply scaled by $\frac{1}{T}$ where $T > 0$ is the temperature \cite{guo2017calibration}. This has the effect that the network becomes less confident for $T>1$ and more confident for $T<1$. The temperature $T$ is selected to maximize the log-likelihood of the validation data. In our setup, temperature scaling only affects the likelihood $p(\hat{y}| x)$ of a translation $\hat{y}$ but not the translation itself as we use the greedy search strategy in eq.\ \eqref{eq:predict}.

\paragraph{Deep Ensembles} DEs \cite{DeepEnsemblesProposed} have been shown to not only provide well-calibrated probabilities, but also to provide superior predictive performance in many settings \cite{ovadia2019can}. DEs are typically implemented by keeping the model architecture and training parameters fixed and simply changing the initialization of the network before training. After fitting the model using $S$ different initializations, we can make predictions by averaging the individual models' probabilities
\begin{align*}
    p_{\text{DE}}(y_i=k \vert x, y_{1:i-1}) = \frac{1}{S} \sum_{s=1}^S  p^{s}(y_i=k \vert x, y_{1:i-1}),
\end{align*}
where $p_{\text{DE}}(y_i=k \vert x, y_{1:i-1})$ denotes the predictive distribution for the DE and $p^s$ denote the $s$'th model in the ensemble.
\section{Experiments}
\label{sec:experiments}
To investigate the proposed methods, we designed and conducted a number of numerical experiments. 

\subsection{Data}
\label{sec:data}
The dataset has been collected from a digital learning platform\footnote{www.writereader.com} and consists of $N = 36,610$ pairs of student and teacher texts, where the teacher text for a given student text is the conventional writing of the student text, i.e.\ the student text modified to have proper spelling, grammar etc. The collected data was filtered to remove sensitive information. We use 80\% of the data for training, 10\% for validation, and 10\% for testing.

The data contains a substantial amount of noise due to unintended use of the learning platform. For example, a student may have written text both in the designated student text field but also in the text field designated for the teacher. Based on a sample of 1,000 pairs, we estimate the percentage of faulty pairs to be around 25\%. To ensure reliable performance estimates, the validation and test dataset have been manually filtered by a teacher to remove faulty pairs. The resulting validation set and test sets contained $N_{\text{val}} = 2,586$ and $N_{\text{test}} = 2,767$.

\paragraph{Data augmentation} We synthetically increase the amount of data we have by simulating students texts and use this in an extra fine-tuning step. We augment conventional children's books provided by Danish publishers with several operations to emulate student text, including word and letter deletions, shortening of words to their initial letter, cutting of word endings, and introduction of common misspellings of letters and bigrams based on the real training data. All operations are applied randomly with heuristically chosen frequencies. This resulted in 279,553 simulated pairs of student and teacher texts.

\subsection{Hyperparameters}
In all experiments, we use the "base" version of BART, which has 6 layers in both the encoder and decoder and a total of 140 million parameters. We use the pre-trained BART model from FAIRSEQ, which has been trained to denoise corrupted text in English \cite{lewis2020bart, ott-etal-2019-fairseq}. We fix the label smoothing parameter to $\epsilon = 0.1$. We further regularize the model with dropout \cite{Dropout} and weight decay \cite{AdamW}. The dropout rate and the amount of weight decay are selected using a grid-search on the validation data, where the median normalized edit distance (see Section \ref{sec:experiment_1}) is used as the selection criterion. Similarly, we train the model until it has converged in terms of the validation median normalized edit distance using the AdamW \cite{AdamW} optimizer with a learning rate of $3\mathrm{e}{-5}$. The text data is encoded with the GPT-2 byte pair encoding \cite{radford2019language}, which has a vocabulary size of $K=50,260$.

\begin{table*}[tp]
    \caption{The baselines and the results of fine-tuning the BART model with the methods described in Section \ref{sec:methods} and the data and metrics described in Section \ref{sec:experiments}. $\pm$ indicates the standard error of the mean. (synth., synthetic, temp., temperature; ED, edit distance; NED, normalized ED; MAE, mean absolute error; FK, Flesch-Kincaid; ECE, expected calibration error; MCE, maximum calibration error).}
    \label{tab:model_table}
    \centering
    \resizebox{\textwidth}{!}{%
    \input{model_table_w_clip.tex}
    }
\end{table*}

\subsection{Reconstructing the teacher texts}
\label{sec:experiment_1}
The purpose of this experiment is to quantify how accurate the teacher texts can be reconstructed from the student texts. 
We assess the quality of a reconstructed text $\hat{y}$ for $x$ by its character-level edit distance (the Levenshtein-distance) to the true teacher text $y$, i.e.\ $\text{ED}(y, \hat{y})$,  that counts the number of deletions, insertions, and substitutions needed to transform $\hat{y}$ into $y$ \cite{levenshtein1966binary}. Since the ED depends on the length of the inputs, we also consider a normalized ED, which is in the interval $[0,1]$:
\begin{align*}
    \text{NED}(y, \hat{y}) = \frac{\text{ED}(y, \hat{y})}{\text{max}(\vert y\vert, \vert\hat{y}\vert )},
\end{align*}
where $\vert y \vert $ and $\vert \hat{y} \vert $ denote the length of the sequences.

We compare the fine-tuned models to three different baseline models. The simplest baseline is the Identity model that simply provides the input text $x$ as the reconstruction $\hat{y}$, i.e.\ $\hat{y}=x$. We also compare against the pre-trained BART model without any fine-tuning, and with reconstructions provided by ChatGPT with the GPT-3.5 Turbo model.

In the first section of Table \ref{tab:model_table}, we report both mean and median EDs for the test set. It is seen that both the Identity model and the pre-trained BART without fine-tuning achieve a mean NED of $0.16$, whereas ChatGPT improves the NED to $0.11$. Fine-tuning of the BART model with the training dataset further reduces the mean NED to $0.09$. The best mean NED of $0.08$ and median NED of $0.01$ was achieved by first fine-tuning with the synthetic student texts and subsequently fine-tuning with the real training data.

\subsection{Robust likelihood}
The next experiment is designed to investigate the benefit of the novel robust likelihood proposed in eq.\ \eqref{eq:rllh}. This likelihood requires the use of an external language model. We employed simple $n$-gram models with $n=2,4,6$, which were estimated using the training data via the KenLM toolkit \cite{heafield-2011-kenlm}. We set $\alpha=0.25$ in the Bernoulli distribution and also regularize the model with dropout and weight decay as with the label smoothed cross-entropy loss.

In the middle section of Table \ref{tab:model_table}, we again report mean and median EDs. We observe that the proposed robust likelihood with a 2-gram language model reduces the mean ED to $3.20$ and the use of a more complex language model (6-gram) further reduces the mean ED to $3.11$ and the median ED and NED to $0$. We do, however, observe that the standard errors of the mean ED and NED for the robust likelihood models overlap with the standard errors for the best model fine-tuned with the smoothed cross-entropy loss.

\subsection{Predicting the linguistic metrics}
\label{sec:experiment_3}
In the third experiment, we investigate how accurately we can estimate the linguistic metrics. We compute the linguistic metrics on the teacher texts $y_n$ and consider those the ground truth. We then evaluate the same metrics using the reconstructed texts $\hat{y}_n$ and compare these to the ground truth.

In this work, we focus on two simple metrics for text complexity and readability, namely the Flesch-Kincaid grade level formula (FK) \cite{kincaid1975derivation} and the readability index LIX \cite{bjornsson1968lasbarhet}, which are given by
\begin{align*}
    \mathrm{FK} &= 0.39\frac{\textrm{\small No. words}}{\textrm{\small No. sentences}} + 11.8 \frac{\textrm{\small No. syllables}}{\textrm{\small No. words}} - 15.59\\
    \mathrm{LIX} &= \frac{\textrm{\small No. words}}{\textrm{\small No. sentences}} + 100\frac{\textrm{\small No. long words}}{\textrm{\small No. words}} \,,
\end{align*}
where long words are defined as words with more than 6 characters. We clip the FK and LIX predictions to the interval $[a,2b]$, where $a$ is the theoretical lower limit of the metric and $b$ is the threshold value for very complex texts. We have that $(a,b)=(-3.4,18)$ and $(a,b)=(0,55)$ for FK and LIX, respectively.

Table \ref{tab:model_table} summarizes the results. We observe that all fine-tuned models achieve comparable performance for LIX, but substantially lower MAEs compared to the baseline models. We also observe the same pattern for the FK metric.

\begin{figure*}[t]
    \centering
    \includegraphics[width=\textwidth]{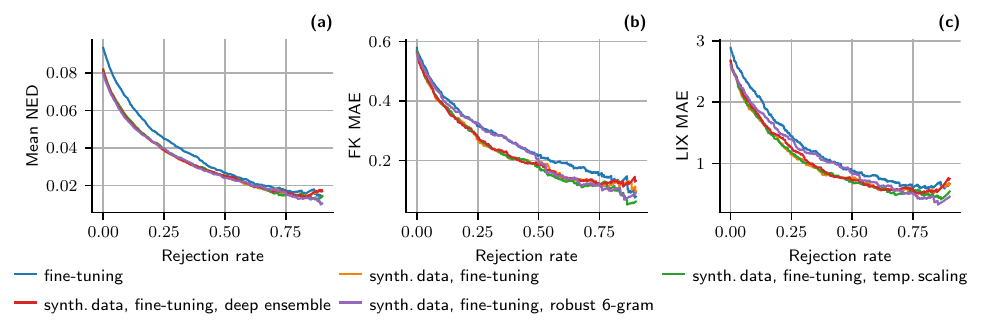}
    \caption{\textbf{(a)} Accuracy-rejection curve for the mean NED and a selection of models from Table \ref{tab:model_table}. \textbf{(b)}-\textbf{(c)} Same curves as in \textbf{(a)} but for the FK MAE and LIX MAE, respectively. The data points in all figures are obtained by varying the rejection threshold for the average log-likelihood $C(\hat{y}| x)$ of a translation $\hat{y}$. "Lower" curves are better.}
    \label{fig:rejection}
 \end{figure*}

\subsection{Calibration and decision-making}
\label{sec:experiment_4}
The purpose of the last experiment is to evaluate and compare the models in terms of calibration and to investigate whether the average log likelihood of the predicted sequences can be used to identify poor predictions. The miscalibration of a model is the difference between the models' confidence and the probability of the model being correct. We quantify the calibration error using both the expected calibration error (ECE) and the maximum calibration error (MCE) \cite{guo2017calibration}. We compute the calibration metrics over all tokens in the test set.

For this experiment, we further compare against a temperature scaled model and a DE constructed using three fine-tuned BART models from random initializations of the model parameters. The last two columns in Table \ref{tab:model_table} summarizes the results.  
It can be observed that all models (except the BART model with pre-training only) perform similar in terms of ECE. Nonetheless, the temperature scaled model is best calibrated, as expected. 

In terms of MCE, the temperature scaled model is, however, not the best calibrated model. The BART model fine-tuned only on the real training data shows the largest MCE, but interestingly, the fine-tuning with synthetic student data greatly reduces the MCE. We also note that the DE method does not lead to improved calibration compared to the single model. However, it does slighly improve the reconstructions w.r.t.\ the mean ED.

Figure \ref{fig:rejection} shows accuracy-rejection curves \cite{nadeem10a} for the mean NED, FK MAE, and the LIX MAE for a selection of models from Table \ref{tab:model_table}. The data points in the plots are obtained by varying the log-likelihood threshold for accepting and rejecting predictions. Figure \ref{fig:rejection} reveals that the average log-likelihood $C(\hat{y}| x)$ is indeed a viable feature for implementing a reject option as all three metrics improves as we reject more translations $\hat{y}$. For example, the mean NED can be reduced from approx. 0.08 to 0.04 if one is willing to reject 25\% of the test observations.

Finally, we see that the rejection curves for all models are quite similar. This suggests that the model calibration does not influence the decision-making abilities of the models, although this result could also be due to the relatively small calibration and performance differences between the models.

\section{Conclusion}
\label{sec:conclusion}
In this work, we have framed the automated feedback on children's early writing as a machine translation problem, where we translate students' early writing into conventional writing. We demonstrated that the conventional writing can be predicted with high accuracy by fine-tuning a pre-trained BART architecture. We also showed that the readability metrics, Flesch-Kincaid and LIX, can be estimated with significantly higher accuracy using the translations compared to the student texts directly. 
Furthermore, as an alternative to the label-smoothed cross-entropy loss function, we proposed a novel robust likelihood to mitigate the effects of noise in the observed data. Our experiments indicated a slightly improved predictive accuracy. Finally, we have shown that the log-likelihood can be used as a criterion for identifying poor translations of the sequence-to-sequence models, inducing a trade-off between accuracy and rejected predictions.

\section*{Acknowledgments}
\label{sec:Acknowledgments}
We acknowledge funding from Innovation Fund Denmark through the InnoBooster program (grant number 2055-00497B). We sincerely thank Janus Madsen and Lasse Sørensen from WriteReader for providing and curating the dataset as well as providing the ChatGPT-based translations of the student texts. We also thank the publishers Alinea and Gyldendal for providing the children's books used to create synthetic data.

\bibliographystyle{abbrvnat}
\bibliography{references}
\end{document}

%% file: model_table_w_clip.tex
\begin{tabular}{l|cc|cc|cc|cc}
\toprule
 & \multicolumn{2}{c}{MEAN} & \multicolumn{2}{c}{MEDIAN} & \multicolumn{2}{c}{MAE} & \multicolumn{2}{c}{} \\
 & ED & NED & ED & NED & FK & LIX & ECE & MCE \\
\midrule
Identity & 5.40$\pm$0.12 & 0.16$\pm$0.00 & 4.00 & 0.12 & 1.36$\pm$0.06 & 5.66$\pm$0.26 &  &  \\
ChatGPT & 5.19$\pm$0.22 & 0.11$\pm$0.00 & 2.00 & 0.05 & 0.78$\pm$0.05 & 3.36$\pm$0.20 &  &  \\
BART (no fine-tuning) & 5.67$\pm$0.14 & 0.16$\pm$0.00 & 4.00 & 0.12 & 1.39$\pm$0.06 & 5.77$\pm$0.26 & 0.19 & 0.29 \\
BART (fine-tuning) & 3.65$\pm$0.14 & 0.09$\pm$0.00 & 1.00 & 0.02 & 0.58$\pm$0.03 & 2.88$\pm$0.14 & 0.04 & 0.65 \\
BART (synth.\ data, fine-tuning) & 3.23$\pm$0.14 & \textbf{0.08}$\pm$0.00 & 1.00 & 0.01 & 0.57$\pm$0.03 & 2.65$\pm$0.14 & 0.04 & \textbf{0.07} \\
\midrule
BART (synth.\ data, fine-tuning, robust 2-gram) & 3.20$\pm$0.13 & \textbf{0.08}$\pm$0.00 & 1.00 & 0.01 & 0.57$\pm$0.03 & 2.65$\pm$0.14 & 0.03 & 0.14 \\
BART (synth.\ data, fine-tuning, robust 4-gram) & \textbf{3.11}$\pm$0.14 & \textbf{0.08}$\pm$0.00 & 1.00 & 0.01 & \textbf{0.55}$\pm$0.03 & 2.61$\pm$0.14 & 0.04 & 0.23 \\
BART (synth.\ data, fine-tuning, robust 6-gram) & \textbf{3.11}$\pm$0.14 & \textbf{0.08}$\pm$0.00 & \textbf{0.00} & \textbf{0.00} & 0.56$\pm$0.03 & \textbf{2.60}$\pm$0.13 & 0.04 & 0.14 \\
\midrule
BART (synth.\ data, fine-tuning, temp.\ scaling) & 3.23$\pm$0.14 & \textbf{0.08}$\pm$0.00 & 1.00 & 0.01 & 0.57$\pm$0.03 & 2.66$\pm$0.14 & \textbf{0.02} & 0.11 \\
BART (synth.\ data, fine-tuning, deep ensemble) & 3.16$\pm$0.14 & \textbf{0.08}$\pm$0.00 & 1.00 & 0.01 & 0.56$\pm$0.03 & 2.68$\pm$0.14 & 0.04 & 0.12 \\
\bottomrule
\end{tabular}